# An Interpretable ML-based Model for Predicting *p-y* Curves of Monopile Foundations in Sand

*Li Biao[a,b], Song Qingkai[a,b], Qi Wengang[a,b*], Gao Fu-Ping[a,b]*

[a] Institute of Mechanics, Chinese Academy of Sciences, Beijing 100190, China
[b] School of Engineering Science, University of Chinese Academy of Sciences, Beijing 100049, China

ABSTRACT.

Predicting the lateral pile response is challenging due to the complexity of pile-soil interactions. Machine learning (ML) techniques have gained considerable attention for their effectiveness in non-linear analysis and prediction. This study develops an interpretable ML-based model for predicting *p-y* curves of monopile foundations. An XGBoost model was trained using a database compiled from existing research. The results demonstrate that the model achieves superior predictive accuracy. Shapley Additive Explanations (SHAP) was employed to enhance interpretability. The SHAP value distributions for each variable demonstrate strong alignment with established theoretical knowledge on factors affecting the lateral response of pile foundations.

## INTRODUCTION

Monopiles are a commonly used foundation type for offshore wind turbines. The lateral bearing capacity of monopile foundations is typically evaluated using the Beam on Nonlinear Winkler Foundation (BNWF) model in engineering design, which represents soil resistance as a series of nonlinear springs that respond to corresponding pile deformations. As a kind of BNWF model, the *p-y* curve method, show in Figure 1, has been extensively employed to quantify the lateral bearing capacity of piles subjected to horizontal loads. This method characterizes the lateral pile-soil interaction along the depth of the soil by defining the relationship between the lateral soil resistance per unit length (*p*) and the lateral deflection (*y*).

Most of the existing *p-y* curve models for sandy soils are based on field tests of small-diameter, flexible piles in dense sand. The *p-y* curve model proposed by Reese et al. (1974) consists of two straight lines connected by a parabolic function. Later, O'Neill and Murchison (1983) simplified the backbone function to a continuous hyperbolic tangent function, which was incorporated into the latest API standards (API, 2014). In this model, the soil resistance *p* at a specific depth *z* due to pile deformation is primarily influenced by the coefficient *A*, ultimate resistance $p_u$, and the initial stiffness coefficient $k_{ini}$. The coefficient *A* is determined from the loading method, while $p_u$ can be calculated based on the friction angle *φ*, relative density $D_r$, effective unit weight *γ'*, and pile diameter *D*. Existing studies have found that the API *p-y* model tends to overestimate the soil reaction *p*. This overestimation arises from the linear relationship between the initial stiffness $E^*_{py}$ and depth *z*, as well as the hyperbolic backbone function used in the model (Georgiadis et al., 1992; Klinkvort, 2012; Wang et al., 2023; Zhu et al., 2016). On the other hand, with the increasing scale of offshore wind turbines, the API *p-y* curve model, originally developed for flexible piles, is gradually becoming insufficient to meet the design requirements for large-diameter rigid piles (Byrne et al., 2015a; Dyson and Randolph, 2001; Thieken et al., 2015; Wang et al., 2022b; Zhang et al., 2024). Some studies have introduced pile diameter *D* as a correction factor to modify the initial stiffness $E^*_{py}$ in the API *p-y* model (Sørensen et al., 2010; Wiemann et al., 2004). Other studies suggest that the initial stiffness $E^*_{py}$ for a given soil layer is directly related to the small-strain elastic modulus $E_s$ and is minimally affected by pile diameter (Ashford and Juirnarongrit, 2003; Wan et al., 2021; Wang et al., 2023). The flexural rigidity $E_pI_p$, which increases with pile diameter, has been found to have no significant effect on the *p-y* curves (Fan and Long, 2005; Suryasentana and Lehane, 2016). The flexural rigidity $E_pI_p$ primarily governs the failure mode of the pile (Poulos and Hull, 1989). In addition, a decrease in the slenderness ratio $L_p / D$ causes the pile foundation to transition from a flexible long pile to a short rigid pile (Byrne et al., 2015b; Chortis et al., 2020; Wang et al., 2021). For short rigid piles undergoing rotational failure, the lateral resistance *p* calculated from sectional moments includes the contribution of side friction. This additional factor can lead to the failure of traditional *p-y* models in accurately evaluating the lateral response of pile foundations.

The findings from previous studies reveal that the API *p-y* curve model is insufficient for accurately analyzing complex pile-soil interactions. Machine learning (ML), which employs computational algorithms to detect and predict statistical patterns in input-output data, presents a promising alternative for addressing these intricate geotechnical challenges. However, the effectiveness and robustness of ML models largely depend on the quantity and quality of data available for supervised training. In geotechnical engineering, such data are often scarce or costly to obtain, posing significant limitations. Moreover, the "black-box" nature of ML models, which lack explicit grounding in physical or mechanical principles, can hinder their interpretability. This lack of transparency raises valid concerns regarding the reliability and practicality of ML-based solutions in real-world design application (Ljung, 2001; Rudin, 2019; Wang et al., 2020). Ouyang et al. (2024) adopt physics-informed neural networks (PINNs) to analyze soil–structure interaction in laterally loaded piles, integrating prior physical information to reduce data needs compared to traditional machine learning methods. In addition to including physical law information described by governing equations in the model, Interpretable Machine Learning (IML) can also provide users with the ability to correctly interpret the prediction results and underlying mechanisms. Kim et al. (2024) developed a scour depth prediction model using eXtreme

Gradient Boosting (XGBoost) and validated the model by comparing the Shapley Additive Explanations (SHAP) distribution of each input feature with existing scour experiments, analyzing the importance of each feature.

Building upon the aforementioned framework, this study introduces an interpretable model for the accurate and reliable prediction of *p-y* curves for pile foundations. The proposed approach begins with training an XGBoost model, optimized through Bayesian tuning, using a dataset of measured *p-y* curves compiled from existing literature. The trained XGBoost model outputs the *p-y* curve at a specified depth, and this process is iteratively repeated to generate p-y curves across varying depths. These curves are subsequently used to back-calculate the lateral response of the pile foundation, which is validated against experimental data. Finally, the Shapley Additive Explanations (SHAP) method is applied to interpret the patterns captured by the model, offering insights into the relationships between input features and the lateral pile response.

## METHODOLOGY

### Database

To achieve accurate prediction results, model training requires a database that represents a comprehensive range of samples. The database utilized in this study comprises 221 *p-y* curves (a total of 2,554 data points) extracted from 19 representative studies on pile lateral responses. Of these, 43 *p-y* curves from three sources were derived using numerically validated models that underwent rigorous verification (Li et al., 2024; Liu et al., 2023). The dataset is partitioned into three subsets: the training set, the validation set, and the testing set. The training set and validation set constitute 70% and 15% of the total dataset, respectively, and are essential for optimizing model weights and hyperparameters during training. The testing set, which comprises the remaining 15%, is employed to evaluate the performance of the trained model. Three curves from Qi et al. (2016), Choo and Kim (2016) and Wang et al. (2021) are excluded from both the training and testing processes and are instead used to evaluate the generalization performance of model.

Fig 1 presents a violin plot with box plots that provide a visual representation of the statistical distribution of feature values, reflecting both their distribution and central tendency. Each feature value is normalized to a range between 0 and 1. The dots in the plot represent the median, and the upper and lower boundaries of the white rectangular box correspond to the first (Q1) and third (Q3) quartiles, displaying the interquartile range (IQR) covering 25% to 75% of the data. The whisker lines extending from the box represent the observations that exceed the third quartile but remain within a specified range, typically defined as 1.5 times the IQR. Data points outside the whiskers may not be accurately predicted by the trained model. Table 1 summarizes the key statistical values of the features. In the violin plot, Kernel Smoothing is applied to generate smoothed density curves for each feature, with different colors indicating varying density levels. Wider regions in the plot indicate higher data concentration, while narrower regions suggest sparsity of data points.

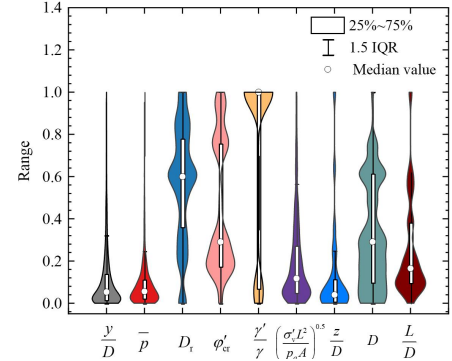

**Fig 1.** Violin plot for the features.

**Table 1.** Data ranges used in developing models.

|  |  | min | max | Q1 | Q3 | 1.5IQR |
|---|---|---|---|---|---|---|
| $D_r$ | - | 0.60 | 1.00 | 0.79 | 0.90 | 1.00 |
| $\varphi'_{cr}$ | ° | 28.50 | 37.10 | 30.00 | 35.02 | 37.10 |
| $\gamma'/\gamma$ | - | 0.48 | 1.00 | 0.52 | 1.00 | 1.00 |
| $D$ | m | 0.24 | 8.00 | 1.00 | 5.00 | 8.00 |
| $\left(\dfrac{\sigma'_v L_p^2}{A p_a}\right)^{0.5}$ | - | 2.50 | 120.14 | 8.25 | 34.55 | 68.92 |
| $z/D$ | - | 0.20 | 6.13 | 1.00 | 3.00 | 5.14 |
| $y/D$ | - | 0.00 | 0.43 | 0.01 | 0.05 | 0.13 |
| $p/\gamma'zD$ | - | 0.00 | 96.84 | 2.15 | 12.15 | 27.11 |

The geometric characteristics of the monopile and the mechanical properties of the sand are the primary factors determining the lateral response of the pile foundation. This database includes various factors that may be related to the lateral bearing capacity of the pile foundation

$$p = f(D_r, \varphi'_{cr}, \gamma', \gamma_w, D, L_p, E_p, z, y) \quad (1)$$

where $\gamma_w$ is the unit weight of water; $E_p$ is the elastic modulus of the pile foundation. For typical offshore wind turbine monopile foundations, made of steel, $E_p$ = 210 GPa; $z$ represents the depth. To further simplify and generalize the application of the model, the following nondimensionalization methods for the feature values are used:

$$\dfrac{p}{\gamma' z D} = f_1\left(D_r, \varphi'_{cr}, \left(\dfrac{\sigma'_v L_p^2}{p_a A}\right)^{0.5}, \dfrac{z}{D}, \dfrac{y}{D}, \dfrac{\gamma'}{\gamma_w + \gamma'}\right) \quad (2)$$

where, $\sigma'_v (= \gamma' z)$ represents the vertical effective stress; $p_a$ is the atmospheric pressure; and $A$ is the area of pile section.

### Performance Evaluation

This study evaluates the predictive performance of the proposed machine learning model using three statistical metrics: Root Mean Square Error (RMSE), Scatter Index (SI), and Pearson Correlation Coefficient (CC). RMSE quantifies the average deviation between the model's predictions and the actual values, reflecting the model's accuracy. A smaller RMSE indicates superior performance, with reduced prediction errors. SI characterizes the residuals for each data point, enabling a precise assessment of model performance. A smaller SI signifies less scatter in the predictions, suggesting better model accuracy. CC measures the linear correlation between predicted and observed values. A CC value closer to 1 (or -1) indicates a stronger linear relationship, emphasizing the consistency between predicted and actual trends. The mathematical definitions of these metrics are provided below to support a more thorough understanding:

$$RMSE = \sqrt{\frac{\sum_{i=1}^{m}(y_i^p - y_i^t)^2}{m}} \tag{3}$$

$$SI = \frac{\sqrt{\frac{1}{m}\sum_{i=1}^{m}(y_i^p - y_i^t)^2}}{\overline{y^t}} \tag{4}$$

$$CC = \frac{\sum_{i=1}^{m}(y_i^t - \overline{y_i^t})(y_i^p - \overline{y_i^p})}{\sqrt{\sum_{i=1}^{m}(y_i^t - \overline{y_i^t})^2}\sqrt{\sum_{i=1}^{m}(y_i^p - \overline{y_i^p})^2}} \tag{5}$$

where $m$ represents the total number of data samples, $y_i^p$ denotes the predicted value for the i-th sample, $y_i^t$ denotes the observed value for the i-th sample, $\overline{y_i^t}$ represents the mean of the observed values, and $\overline{y_i^p}$ indicates the mean of the predicted values.

**Bayesian Optimization**

Bayesian Optimization (BO) is an effective technique for optimizing model hyperparameters in machine learning. Unlike blind exploration methods such as grid search or random search, BO directs the optimization process by constructing a probabilistic model that represents the relationship between hyperparameters and the objective function. Typically, BO uses a surrogate probabilistic model, such as Gaussian Process Regression (GPR), to quantify prediction uncertainty. GPR not only predicts the objective function values but also estimates the uncertainty associated with these predictions.

The optimization process begins by training the model on an initial set of hyperparameter combinations and recording the corresponding performance metrics. Based on these results, BO constructs a GPR model, assuming that each objective function value (data point) follows a normal distribution. As new hyperparameter combinations and their corresponding objective function values are evaluated, the surrogate model is updated using Bayes' theorem. This iterative process refines the model's understanding of the objective function, progressively enhancing its approximation of the true structure.

To select the next hyperparameter combination for evaluation, BO employs an acquisition function. This function balances two competing objectives: exploration, which tests untried regions of the hyperparameter space, and exploitation, which focuses on refining promising regions identified by the current model. Exploration prevents the algorithm from becoming trapped in local optima, while exploitation fine-tunes the search around high-performing areas. Through the iterative process of updating the surrogate model and evaluating new hyperparameter combinations, BO converges toward the optimal hyperparameter set or halts once a predefined computational budget is reached.

The process can be mathematically represented by

$$X^+ = \arg\max_{X \in A}(P(X)) \tag{6}$$

where the acquisition function determines the next candidate hyperparameter combination based on the surrogate model's predictions. where $A$ represents the search domain of $X$; $P$ denotes the objective function; $X$ and $X^+$ correspond to the variable to be optimized and the optimal variable, respectively. Table 2 summarizes the search ranges for the hyperparameters of the XGBoost model. Additional information about BO can be found in Eini et al. (2024); Wu et al. (2019).

Table 2. Search space of hyperparameters for the XGBoost and GPR model.

| Model | Hyperparameters | Search Space |
|---|---|---|
| XGBoost | n_estimators | 600-1000 |
| | max_depth | 3-10 |
| | learning_rate | 0.01-03 |
| | subsample | 0.5-1 |
| | min_child_weight | 1-10 |
| GPR | length_scale | 0.001-100 |
| | noise_level | 0.0001-0.1 |

**XGBoost (eXtreme Gradient Boositng) model**

The core concept of XGBoost is based on constructing multiple trees, where each new tree is designed to correct the errors made by the previous trees, thereby progressively improving the model's performance (Chen and Guestrin, 2016). For a given dataset $\mathbb{N} = \{(x_i, y_i)\}(|\mathbb{N}| = n, x_i \in \mathbb{R}^m, y_i \in \mathbb{R})$ with $n$ examples and $m$ features, a tree ensemble model uses $K$ additive functions to predict the output

$$\hat{y}_i = \phi(x_i) = \sum_{k=1}^{K} f_k(x_i), \quad f_k \in \Re \tag{7}$$

where $\Re = \{f(x) = w_{q(x)}\}(q_T : \mathbb{R}^m \to T, w \in \mathbb{R}^T)$ is the space of regression trees. Here $q_T$ represents the structure of each tree that maps an example to the corresponding leaf index. $T$ is the number of leaves in the tree. Each $f_k$ corresponds to an independent tree structure $q_T$ and leaf weights $w$. Unlike decision trees, each regression tree contains a continuous score on each of the leaf, we use $w_i$ to represent score on $i$-th leaf. The objective function $L$ consists of a loss function $l$ and a regularization term $\Omega$

$$L = \sum_i l(\hat{y}_i, y_i) + \sum_k \Omega(f_k) \tag{8}$$

where $l$ is a loss function that measures the difference between the prediction $\hat{y}_i$ and the target $y_i$. The regularization term $\Omega$ controls the complexity of the model through the bias-variance tradeoff, maintaining simplicity and predictive accuracy, which helps prevent overfitting (Kardani et al., 2020).

Fig. 2 illustrates the process of developing the Optimal XGBoost models to determine the *p-y* curves of a monopile. A series of feature values from a target *p-y* curve is input into the optimized XGBoost model to generate a set of predicted points.

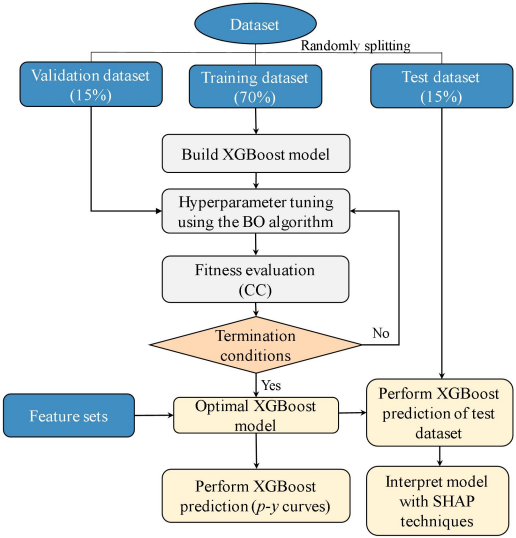

Fig.2. Process flow of the XGB model in estimating the *p-y* curves of monopiles.

### SHAP(Shapley Additive Explanations)

SHAP (Shapley Additive Explanations) is a machine learning interpretability technique that quantifies the contribution of each feature to the model's predictions. It achieves this by comparing the model's output with and without the inclusion of a specific feature, thereby highlighting its importance in the prediction process (Shapley, 1953). One of the key advantages of SHAP is its versatility in explaining machine learning models, whether they are linear, tree-based, or deep learning models. Furthermore, SHAP offers a consistent framework for analyzing feature significance, enhancing the model's transparency, interpretability, and trustworthiness (Kardani et al., 2020; Kim and Lee, 2024; Kim et al., 2024). To build interpretable models, SHAP employs an additive feature attribution method, which expresses the model's output as the linear sum of the input features (Eini et al., 2024; Lundberg et al., 2018). In this method, each input feature is assigned a SHAP value

$$v_i = \sum_{S \subseteq M} \frac{|S|!(m-|S|-1)!}{m!} \left[ \tau(S \cup \{i\}) - \tau(S) \right]$$
(9)

where $v_i$ the Shapley value of feature $i$, which represents the contribution of the $i$-th feature to the model's output; $M$ represents all input parameters；$m$ is the total number of features; $S$ is a subset of $M$; $\tau(S \cup \{i\})$ and $\tau(S)$ represent the contributions of the model with and without the inclusion of the $i$-th parameter, respectively (Eini et al., 2024; Khattak et al., 2022). Once the Shapley values for all input parameters are obtained, local interpretations can be made for each observation, and global explanations can be provided based on the average SHAP values of each parameter. In this study, SHAP values are used to analyze the features of the trained XGBoost model. The SHAP values can be applied to interpret the model and extract the significance of each parameter. These values can be either positive or negative, indicating the direction of the parameter's influence on the model's predictions (Eini et al., 2024; Feng et al., 2021; Kim et al., 2024).

## RESULTS AND DISCUSSION

### Performance of the XGBoost model in predicting *p-y* curves

Fig. 3 illustrates the correspondence between observed and predicted values using the Extreme Gradient Boosting (XGBoost) algorithm. The proposed XGBoost model demonstrates exceptional performance on the training set, with a Root Mean Square Error (RMSE) of 0.172, a Scatter Index (SI) of 0.019—indicating highly stable predictions—and a Correlation Coefficient (CC) of 0.999, signifying near-perfect alignment.

On the test set, the model also performs well, achieving an RMSE of 1.034. Although this value is slightly higher than the training set RMSE, it remains relatively low. Additionally, the SI is 0.114, indicating minimal scatter, and the CC is 0.990, reflecting a strong linear relationship between predicted and observed values. These results collectively affirm the XGBoost model's high accuracy and robust generalization capability for the dataset in this study.

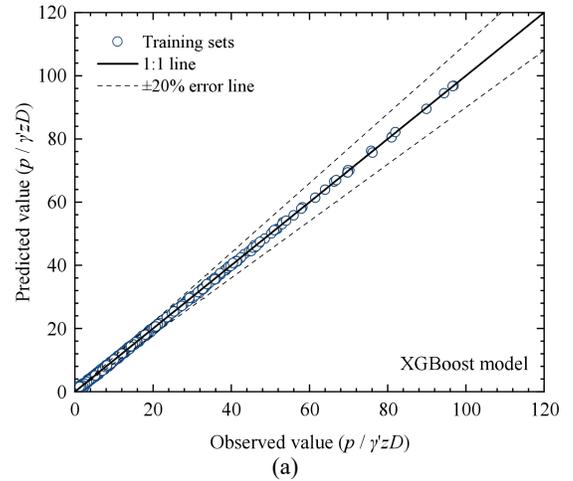

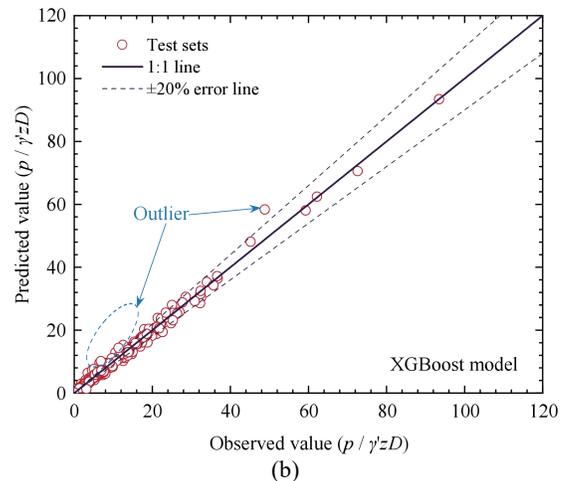

Fig.3 Comparison of $p/\gamma'zD$ between the observed and predicted values of XGBoost model for the (a) training and (b) test datasets.

Three curves from Qi et al. (2016), Choo and Kim (2016) and Wang et al. (2021) are excluded from both the training and testing processes to evaluate the generalization performance of XGBoost model. The comparison results in Fig.4 show that the output points from the XGBoost model are closely distributed around the corresponding p-y curve. However, these scatter points cannot be directly connected to form a smooth p-y curve. This phenomenon occurs because the sampling ranges and densities of the p-y curves from different sources in the training set may not be entirely consistent. Therefore, it is necessary to apply a smoother, such as LOESS (Locally Weighted Regression), to generate a smooth p-y curve.

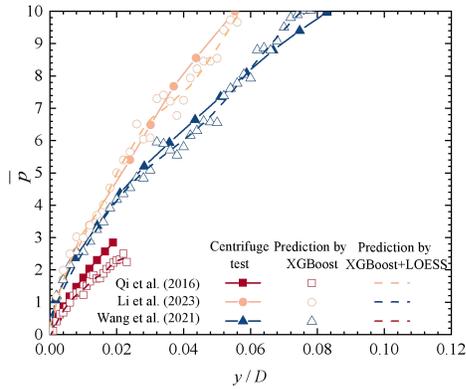

Fig.4 Comparison of p-y curves between the observed data, predicted data and smoothed curves.

**Comparison with centrifuge tests and field tests**

Using the XGBoost model, a series of p-y curves along the pile can be generated, which can subsequently be used to back-calculate the force-displacement and moment-depth curves of the monopile. This approach validates the generalization capability of the predictive model. The p-y data predicted by the XGBoost model can also be utilized in the Oasys ALP program (referred to as ALP) to compute the lateral pile response. Table 3 summarizes the key parameters and feature value ranges used for validation in the studies by Qi et al. (2016), Reese et al. (1974) and Choo and Kim (2016). As illustrated in Figures 5 and 6, the lateral pile response back-calculated from the predicted p-y curves closely aligns with the centrifuge test results, demonstrating that the trained XGBoost model is effective in assessing the lateral load-bearing capacity of piles.

**Table 3**. Prediction parameters used in the comparasion with tests

| Parameters | Qi et al. (2016) | Reese et al. (1974) | Choo et al. (2016) |
|---|---|---|---|
| $D$ | 2.75 m | 0.61 m | 6.00 m |
| $L_p / D$ | 11.36 | 34.43 | 5.17 |
| $D_r$ | 0.68 | 0.65 | 0.86 |
| $\varphi'_{cr}$ | 35° | 28.5° | 35.2° |
| $\dfrac{z}{D}$ | 0 ~ 11.36 | 0 ~ 34.43 | 0 ~ 5.17 |
| $\dfrac{\gamma'}{\gamma_w + \gamma'}$ | 0.51 | 0.48 | 0.51 |

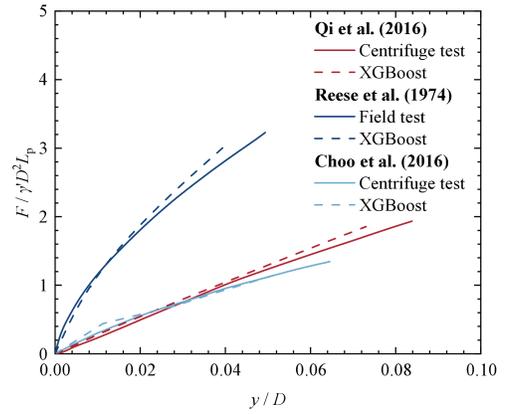

Fig.5. Comparison of force-displacement curves obtained from centrifuge tests and back-analysis with predicted p-y curves.

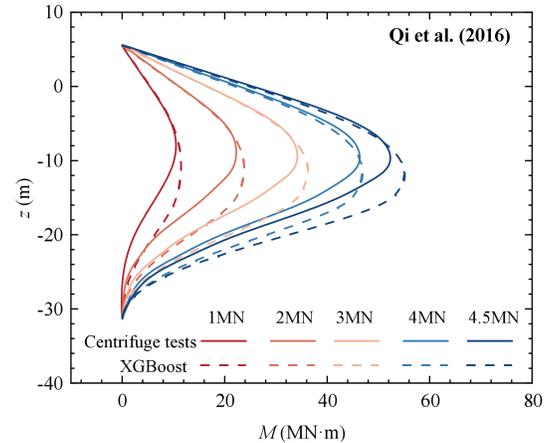

(a)

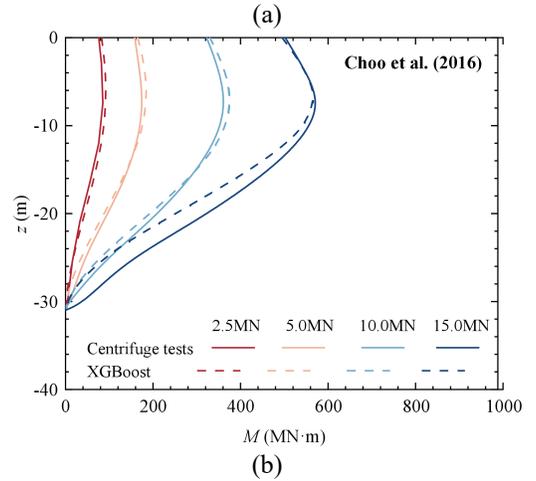

(b)

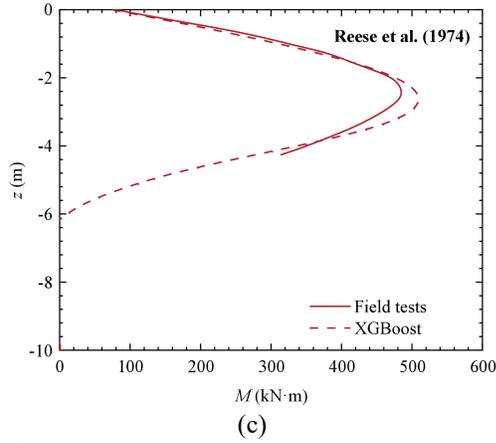

(c)

Fig.6. Comparison of moment-depth curves obtained from centrifuge tests and back-analysis with predicted *p-y* curves.

**MODEL INTERPRETATION**

The XGBoost model has demonstrated its capability to provide accurate predictions on both the training and test datasets. However, the complex internal structure of such models presents a significant challenge in comprehending their prediction mechanisms. SHAP (SHapley Additive exPlanations), as a robust analytical tool, provides valuable insights into the inner workings of complex machine learning models like XGBoost, which involve numerous parameters (Kim et al., 2024).

This section utilizes the SHAP method to comprehensively explain the principles underlying the XGBoost model's predictions. By analyzing the interactions between individual features and model outputs, the study seeks to bridge the gap between the model's construction and the associated physical mechanisms. In particular, the analysis emphasizes the influence of the "diameter effect" on evaluating lateral pile responses.

Global interpretation aims to summarize the SHAP values of input parameters within the dataset and provide an estimation of each input feature's relative influence on the dependent variable. Figure 10(a) illustrates the importance analysis results for each input parameter in the prediction of dimensionless soil resistance $\bar{p}$. Specifically, it presents the global interpretation based on the average absolute Shapley values corresponding to each feature. Among the six features, the average SHAP value of $y/D$ is the highest, indicating its dominant influence on the predictive model for soil resistance. In contrast, the ratio of effective unit weight to total unit weight has the least impact. The influence of the remaining features on the model is relatively similar.

Figure 10(b) illustrates the distribution of feature values and their corresponding SHAP values. Negative SHAP values indicate a negative (-) effect, while positive SHAP values represent a positive (+) effect for the prediction. Red points correspond to higher values for each feature value, while blue points indicate lower values. Increasing deformation shifts its impact on the predictive model from negative to positive. SHAP values increase with both relative density and critical friction angle. The feature $\left(\dfrac{\sigma'_v L_p^2}{A p_a}\right)^{0.5}$ is primarily used to assess the deformation patterns of the pile foundation. It consists of two components: one representing soil stiffness ($\dfrac{\sigma'_v}{p_a}$) and the other related to deformation ($\dfrac{L_p^2}{A}$). The initial elastic modulus of homogeneous sand is considered to depend on the confining pressure determined by self-weight stress (Achmus et al., 2009; Byrne et al., 1987; Hardin, 1987; Hryciw and Thomann, 1993; Li et al., 2024). Existing studies indicate that the initial lateral stiffness of the *p-y* curve in any given soil layer is directly correlated with the small-strain elastic modulus of the soil (Randolph, 1981; Wan et al., 2021; Wang et al., 2021). Variations in the slenderness ratio influence the failure mechanism of laterally loaded piles (Thieken et al., 2015; Wang et al., 2021). Large-diameter short rigid piles experience axial side friction during rotational failure. Notably, the *p-y* curves commonly used in researches are derived by double differentiation of the measured bending moment profile in depth *z*

$$p = \dfrac{d^2 M}{dz^2} \tag{10}$$

$$\dfrac{M}{E_p I_p} = \dfrac{1}{\rho} \tag{11}$$

The bending moment *M* at a specific depth is typically derived from the curvature *ρ*, using strain gauges placed diametrically opposite on the tension and compression sides of the pile, as expressed in Eq. (11). Here, $E_p I_p$ represents the flexural rigidity of the pile section. Since the measured strain represents a point value synthesizing all forces acting on the pile, the product of axial friction *f* and the radius *R* contributes to the measured sectional moment (i.e., tensile and compressive strains). Consequently, these effects are implicitly incorporated into the experimentally and numerically derived *p-y* curves (Li et al., 2024; Wang et al., 2022a). As pile length $L_p$ and the vertical effective stress of the soil $\sigma'_v$ decrease, Fig.10 a diminishing influence of this feature value, accompanied by a reduction in the predicted value. This trend aligns with the observation that side friction contributes less to the reaction force resisting pile deformation under these conditions. In the study, the ratio of effective unit weight to total unit weight has a minimal impact on the prediction results. Klinkvort (2012) demonstrated that tests conducted using dry sand provided consistent results compared to tests conducted in saturated sand at the same effective stress level. Suryasentana and Lehane (2016) suggested that the *p-y* curves below the groundwater table are slightly stiffer than those in completely unsaturated conditions. The negligible effect may be attributed to the fact that most of the data originates from centrifuge tests, where the water depth above the soil layer is relatively shallow.

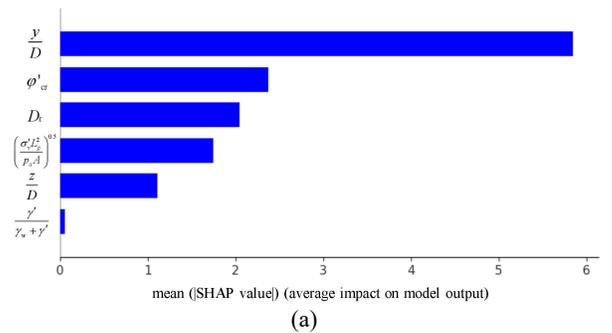

(a)

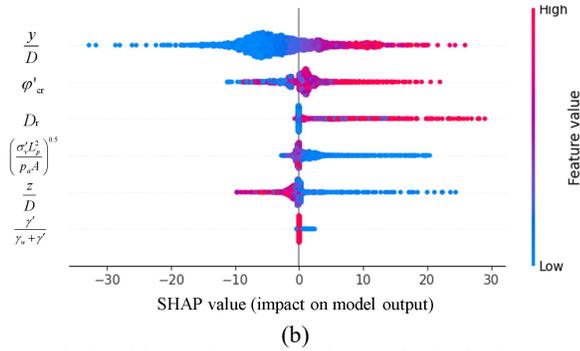

(b)

Fig. 7. Analysis of feature importance for $\bar{p}$ obtained using the trained XGBoost model:(a) SHAP feature importance and (b) summary plot (feature effects).

CONCLUSIONS

This study proposes a method for predicting *p-y* curves of pile beam models using an XGBoost model. A dataset comprising 221 *p-y* curves (totaling 2,554 points) was collected from 19 representative studies on lateral pile response to train the predictive model. The reference dataset includes pile failure modes for both flexible piles and short rigid piles. The trained model can directly generate *p-y* curves, facilitating the evaluation of the lateral bearing capacity of piles.

(1) The trained XGBoost model demonstrates high accuracy in predicting soil resistance *p* under different deformations, with the test dataset showing a Correlation Coefficient (CC) of 0.956.

(2). Combining the XGBoost model with the LOESS smoother enables accurate prediction of the *p-y* curves for pile beam models, demonstrating strong generalization capability. This approach can be utilized for the preliminary assessment of the lateral bearing capacity of offshore wind turbine monopile foundations.

(3). The interpretation of the predictive model indicates that deformation is the primary factor influencing the soil resistance $\bar{p}$. The ratio of the effective unit weight to the total unit weight of sand has a minimal impact on the prediction of $\bar{p}$. The SHAP value variations of the remaining features similarly align with the physical mechanisms of pile deformation. This alignment explains the model's strong generalization capability.

(4) The model interpretation reveals that the dimensionless parameter $\left(\frac{\sigma'_v L_p^2}{A p_a}\right)^{0.5}$ effectively captures the transition in pile deformation modes from slender pile failure to short rigid pile failure. It also reflects the contribution of side friction to the sectional bending moment *M* during this transition.

ACKNOWLEDGEMENTS

This work was funded by the National Natural Science Foundation of China (Grant Nos. 11972036) and the Youth Innovation Promotion Association CAS (Grant No. 2021018).